\documentclass[accepted]{melba}

\usepackage{mwe} 

\usepackage{amsmath,amsfonts}

\usepackage{physics}
\usepackage{multirow}

\usepackage{algorithm}
\usepackage{algorithmic}

\usepackage{dblfloatfix}

\usepackage{balance}

\usepackage[table]{xcolor}

\usepackage{xcolor}
\definecolor{mylightblue}{RGB}{133, 193, 233}
\definecolor{mycoral}{RGB}{236, 112, 99}

\melbaid{2026:020}  
\doi{10.59275/j.melba.2026-d8c8}
\melbaauthors{Bhattacharya, Singh, Jain and Prasanna}  
\email{moinak.bhattacharya@stonybrook.edu}
\volume{2026}
\firstpageno{411}  
\melbayear{2026}  
\datesubmitted{2025-08-01}  
\datepublished{2026-02-02}  

\ShortHeadings{GazeLT}{Bhattacharya, Singh, Jain and Prasanna}

\title{GazeLT: Visual attention–guided long-tailed disease classification in chest radiographs}

\author{
	\firstname Moinak \surname Bhattacharya\aff{1},
	\firstname Gagandeep \surname Singh\aff{2},
    \firstname Shubham \surname Jain\aff{3},
    \firstname Prateek \surname Prasanna\aff{1},
}
\affiliations{
	\num 1 \addr Department of Biomedical Informatics, Stony Brook University, NY, US, 11790 \\
	\num 2 \addr Department of Radiology, Columbia University, NY, US, 10027 \\
	\num 3 \addr Department of Computer Science, Stony Brook University, NY, US, 11790
}

\abstract{
	In this work, we present \textit{GazeLT}, a \textit{human visual attention integration-disintegration approach} for long-tailed disease classification. A radiologist's eye gaze has distinct patterns that capture both fine-grained and coarser level disease related information. While interpreting an image, a radiologist's attention varies throughout the duration; it is critical to incorporate this into a deep learning framework to improve automated image interpretation. Another important aspect of visual attention is that apart from looking at major/obvious disease patterns, experts also look at minor/incidental findings (few of these constituting long-tailed classes) during the course of image interpretation. \textit{GazeLT} harnesses the temporal aspect of the visual search process, via an integration and disintegration mechanism, to improve long-tailed disease classification. We show the efficacy of \textit{GazeLT} on two publicly available datasets for long-tailed disease classification, namely the NIH-CXR-LT (n=89237) and the MIMIC-CXR-LT (n=111898) datasets. GazeLT outperforms the best long-tailed loss by 4.1\% and the visual attention-based baseline by 21.7\% in average accuracy metrics for these datasets.
	Our code is available at~\url{https://github.com/lordmoinak1/gazelt}.}


\begin{document}

\twocolumn[\maketitle]

\begin{figure*}[h]
    \centering
    \includegraphics[width=1\linewidth]{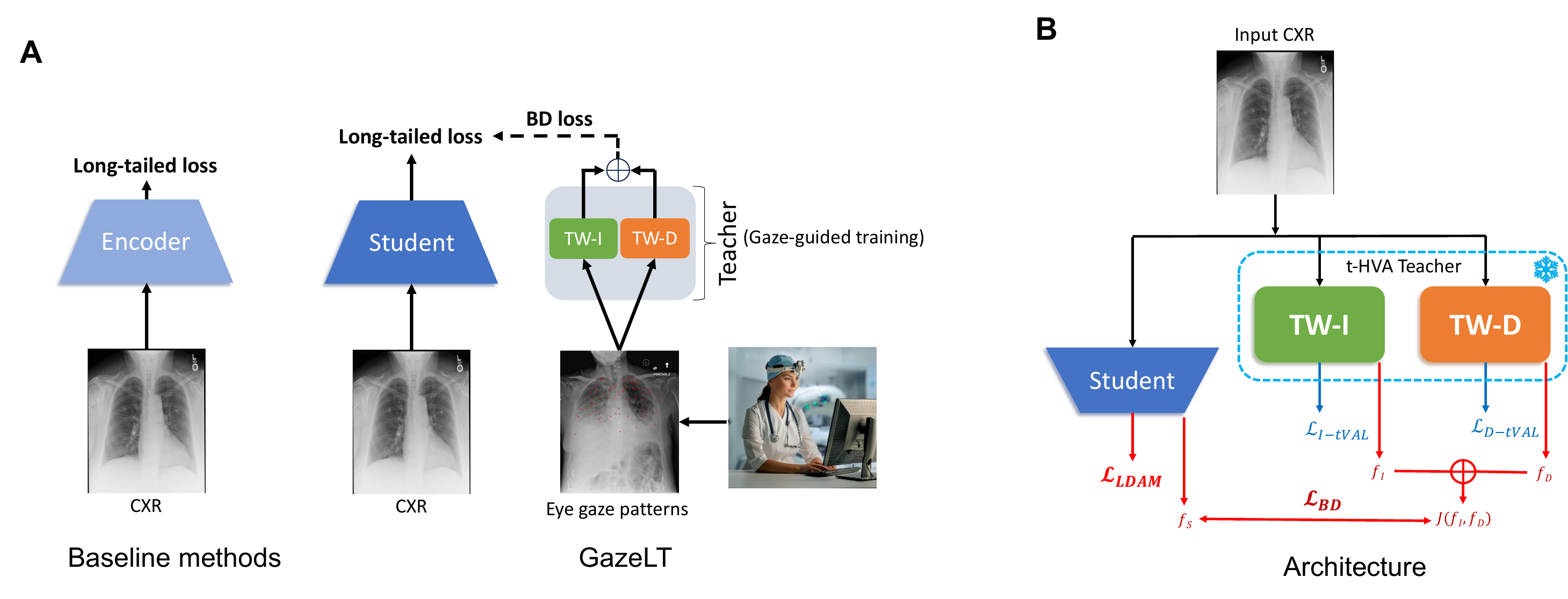} 
    \caption{A. Baseline methods use a vision encoder with long-tailed losses $\mathcal{L}_{LT}$ whereas GazeLT uses a teacher-student framework where the teacher is trained with radiologists’ visual attention patterns and then the student is trained by distilling the knowledge from the teacher using $\mathcal{L}_{BD}$ and the long-tailed loss $\mathcal{L}_{LT}$, B. GazeLT is a teacher-student framework. The t-HVA teacher has two components, TW-I and TW-D. The TW-I is trained using $\mathcal{L}_{I-tVAL}$ and the TW-D is trained using $\mathcal{L}_{D-tVAL}$. During student training, the $f_I$ and $f_D$ from t-HVA are fused and distilled to the student using $\mathcal{L}_{BD}$.
}
\label{fig:main}
\end{figure*}
\begin{figure*}[h]
    \centering
    \includegraphics[width=0.8\linewidth]{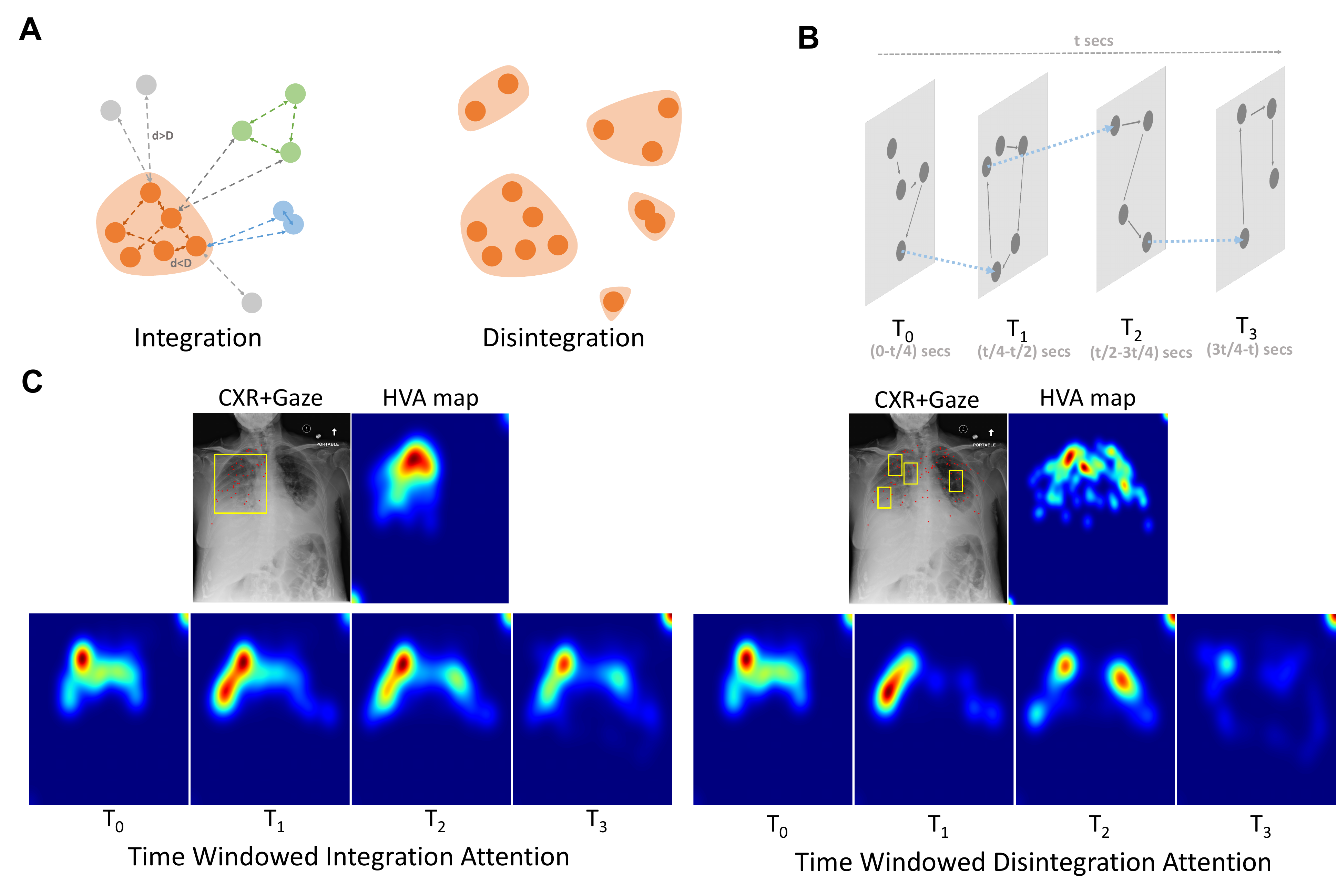} 
    \caption{A. Integration and disintegration of eye gaze points, with orange clusters representing distinct attention calculation mechanisms, B. Partitioning of eye gaze points into specific time intervals for temporal attention modeling, C. Time Windowed HVA maps. The HVA maps at different timepoints, T0, T1, T2 and T3, for Integration and Disintegration are shown.
}
\label{fig:eyegaze}
\end{figure*}
\begin{figure*}[h]
    \centering
    \includegraphics[width=0.9\linewidth]{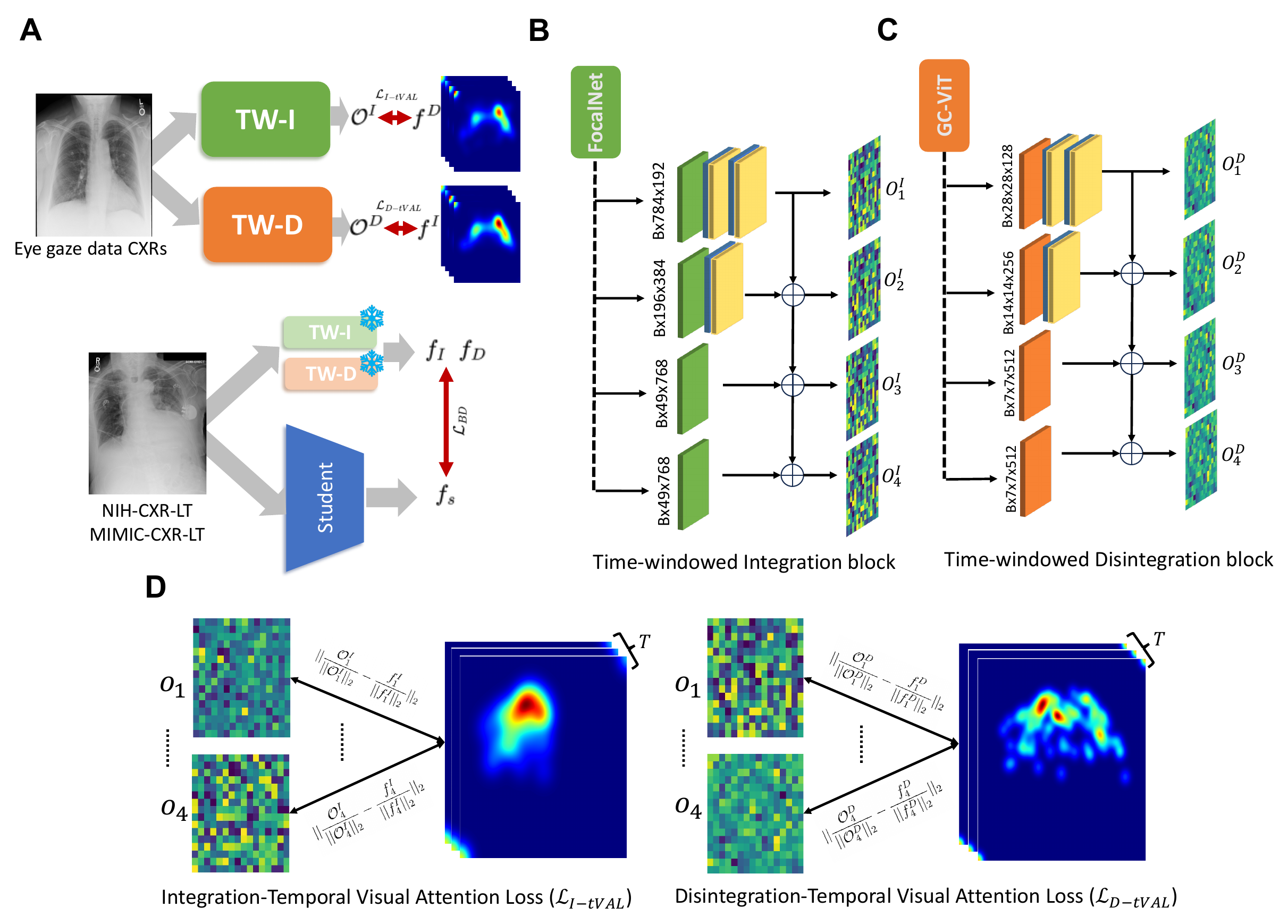} 
    \caption{A. Overall training pipeline illustrating the two-stage process: (i) the teacher network is pretrained using raw gaze points segmented into $n$ time windows to encode temporal visual attention via TW-I and TW-D modules; (ii) the student model is trained with frozen teacher supervision through feature-level distillation. B. TW-I block is a FocalNet transformer architecture from which the integration feature outputs $\mathcal{O}^I_i$ are obtained using a decoder, C. TW-I block is a GC-ViT transformer architecture from which the disintegration feature outputs $\mathcal{O}^D_i$ is obtained using a decoder. D. Time Windowed Visual Attention Loss (tVAL). The loss for the TW-I block, $\mathcal{L}_{I-tVAL}$, and the loss for the TW-D block, $\mathcal{L}_{D-tVAL}$.
}
\label{fig:architecture}
\end{figure*}
\section{Introduction}

	Eye gaze patterns are important in understanding the visuo-cognitive process underlying a radiologist’s interpretation of medical images (\cite{van2017visual}). Visual patterns can be attributed to maintaining, manipulating, and retaining information based on past and current observations (\cite{logie2003spatial}). This information can range from coarse trivial attention to fine-grained non-trivial attention patterns. Coarse attention patterns may arise due to non-salient abnormalities, often not relevant to the diagnosis of any particular disease whereas fine-grained patterns are associated with salient abnormalities that contain information of specific disease patterns and location. Gaze patterns are direct implications of the perception of a salient object in the search view based on the object's physical appearance and location (\cite{jian2021visual}). The temporal aspect of this search becomes critical as, with time, more information gets stored in memory and that directly affects the future gaze patterns (\cite{drew2017one}). As a simple example, in a find-an-object puzzle/game, a person starts looking from the top left corner and after a detailed search concludes that the object-of-interest is not in that particular region; in the future, the probability of searching in that region is lower than searching in other regions. Again, before moving to any other region for further fine-grained search, the person prefers to have a global view of the entire region, and based on random understanding, the next region is selected. This approach of global random search is akin to disintegration attention (\cite{dieter2011understanding}) and the fine-grained search can be compared to the integration attention (\cite{ferreira2014attentional}) in human psychology. We present a method to incorporate the integration and disintegration attentions into a deep learning (DL) framework for better-informed and more interpretable decision-making. This is important as DL models often tend to focus on disease-irrelevant features for disease predictions. Radiologists' eye gaze patterns, collected during diagnosis, can be used to better guide a DL model to focus on disease-relevant regions.\\
Improving the interpretability of DL models in medicine has been an active area of research (\cite{salahuddin2022transparency}). DL models in radiology often tend to learn shortcuts (\cite{ma2023eye}) or look at irrelevant regions while making predictions (\cite{zhang2023diagnostic}). One approach to address this involves the incorporation of clinical experts' visual search patterns into the model (\cite{bhattacharya2022gazeradar,bhattacharya2022radiotransformer,bhattacharya2024gazediff, bhattacharya2024radgazegen}). This \textit{clinician-in-the-loop} strategy is gaining prominence. Recent works show that DL models perform better when clinicians' feedback is incorporated into the decision-making pipeline (\cite{feng2022clinical}). In this work, we propose a novel strategy for integrating clinical experts' visual patterns into DL models. More specifically, our model is explicitly trained with the integration and disintegration attention patterns of radiologists to make a diagnosis. This helps the model make more clinically accurate predictions. We demonstrate the efficacy of long-tailed classification tasks in pulmonary disease classification by integrating the eye gaze patterns of radiologists.\\
Long-tailed classification in medical imaging is a challenging problem. This challenge arises from the disparity in prevalence between common diseases, which are relatively limited in number and easily observable in images such as chest radiographs, and rare diseases, which exhibit a considerably higher variety (\cite{zhou2021review,paul2021generalized}). For example, in most of the chest radiography datasets, diseases like Pneumothorax, Atelectasis, Infiltration, etc., are common pathologies present in $>60\%$ of the total patients, whereas diseases like Pneumomediastinum, Subcutaneous Emphysema, Pneumoperitoneum, etc., are rare pathologies present in $<1\%$ of the total patients. The common disease types are called head classes and the rare disease types are called tail classes (\cite{zhang2023deep}). DL models have been shown to perform better in discriminating the head classes and exhibit poor performance in identifying/discriminating tail classes (\cite{holste2022long}). Recently, several methods have been proposed in the fields of imbalanced learning (\cite{zhou2021review}), and few-shot learning (\cite{feng2021interactive}) to deal with long-tailed classification. However, none of these use clinicians' feedback, in any form, to improve long-tailed learning. We hypothesize that, for long-tailed classification, experts' visual patterns are of prime importance. Radiologists, while reading an image, fixate on obvious abnormalities which are the head classes in the datasets, and also scan through other incidental findings, some of which may correspond to the tail classes (Figure \ref{fig:radint}).\\ 
Our rationale for decomposing the ‘static’ nature of attention as used in previous studies (\cite{stember2020integrating,bhattacharya2022gazeradar,bhattacharya2022radiotransformer,bhattacharya2024gazediff,bhattacharya2024radgazegen}) to a ‘temporally resolved’ form stems from this dynamic nature of viewing patterns of radiologists during diagnosis. This motivates us to compartmentalize the gaze patterns into windows based on durations.  We hypothesize that the gaze patterns in these individual windows fixate, partially or wholly, on individual disease-relevant regions. These regions have context-rich disease information of the individual head, medium, and tail classes. Hence, the integration of these time-windowed human visual attentions into a deep learning framework can facilitate long-tailed learning. In this work, we present GazeLT, a novel temporal visual attention-guided teacher-student deep learning framework for long-tailed classification (Figure \ref{fig:main}).\\
To the best of our knowledge, no other work has investigated the impact of experts' visual attention in long-tailed learning. Our work is based on the premise that temporal visual attention of radiologists can infuse auxiliary diagnostic information into a DL framework to help improve long-tailed classification. Previous works have focused on the static nature of human visual attention (\cite{stember2020integrating,bhattacharya2022gazeradar,bhattacharya2022radiotransformer}). We argue that more detailed and contextual clinical diagnostic information can be retrieved if we decompose this ‘static' nature of radiologists' attention computation into a ‘temporally resolved’ form. The main contributions of this paper are as follows: 

\begin{itemize}
    \item We harness the temporal aspect of the visual search process by developing a novel Time Windowed Integration and Disintegration approach to capture the fine-grained and coarse attention patterns. We propose novel Visual Attention Loss (VAL) functions that help the deep learning model to learn the visual attention patterns,
    \item We propose a teacher-student framework where the teacher is trained with time-windowed visual attention and distills the features to a student that is finetuned for downstream long-tailed classification, and 
    \item We show the efficacy of our method on two medical long-tailed classification benchmarks, NIH-CXR-LT and MIMIC-CXR-LT.
\end{itemize}
We observe that radiologists’ eye gaze patterns offer valuable insights into disease characteristics, particularly within the long-tail distribution of diagnostic categories. Leveraging these patterns in machine learning models presents a promising strategy for addressing the challenges of long-tailed disease classification. To the best of our knowledge, no prior work has incorporated radiologists’ gaze behavior to specifically mitigate the long-tail classification problem. Our proposed approach demonstrates improved performance on publicly available datasets for long-tailed classification of pulmonary diseases.

\section{Related Works}
\subsection{Eye gaze in medical image analysis}
Eye gaze tracking has emerged as a powerful tool in various medical applications, such as disease classification (\cite{bhattacharya2022gazeradar,bhattacharya2022radiotransformer}) and segmentation (\cite{wang2023gazesam}), across multiple modalities including radiology (\cite{stember2020integrating}), pathology (\cite{sudin2022digital}), retinopathy (\cite{clark2019potential}), and ECG analysis (\cite{sqalli2022understanding,tahri2021interpretation}). 
The integration of eye gaze in general computer vision tasks has shown success in several downstream tasks such as object detection (\cite{smith2013gaze,cho2021human}), image segmentation (\cite{amrouche2018activity,james2007eye,shi2017gaze}), action localization (\cite{shapovalova2013action,mathe2014actions}), and activity recognition (\cite{courtemanche2011activity,min2021integrating}). Building on these advancements, recent research has focused on utilizing gaze patterns from medical professionals, such as radiologists and pathologists, to improve the diagnostic performance of ML models and reduce diagnostic errors (\cite{roshan2023eye}). Specifically, teacher-student knowledge distillation models have demonstrated the potential of leveraging gaze data to improve disease diagnosis (\cite{bhattacharya2022gazeradar,bhattacharya2022radiotransformer}). Eye gaze patterns have also been effectively integrated with deep learning frameworks, including graph neural networks (\cite{wang2024gazegnn}), vision transformers (\cite{ma2023eye,bhattacharya2022radiotransformer}), etc. In recent advances, eye gaze data have been used in medical image generation (\cite{bhattacharya2024gazediff, bhattacharya2024radgazegen}). However, the use of radiologists' eye gaze patterns to enhance long-tailed classification in deep learning remains underexplored. These patterns, which highlight rare and incidental findings, can guide models to better predict tail classes. We propose a method that leverages gaze information to improve long-tailed learning performance.
\subsection{Long-tailed disease classification}
Long-tailed learning has emerged as a critical challenge in medical imaging due to the inherent class imbalance in clinical datasets. Rare pathologies are often underrepresented compared to common conditions, leading to performance degradation in standard deep learning models trained with conventional loss functions and sampling strategies. While long-tailed learning has been extensively studied in natural image domains using techniques such as re-sampling (\cite{wang2020devil,estabrooks2004multiple,zhang2021learning}) and re-weighting (\cite{ren2020balanced,elkan2001foundations,jamal2020rethinking,tan2021equalization,tan2020equalization}), these approaches remained largely unexplored in medical imaging until recent advancements. In medical imaging, several studies have begun addressing class imbalance through modifications to loss functions. For example, a unified focal loss was proposed to tackle class imbalance in medical image segmentation (\cite{yeung2022unified}). Weighted class-balanced loss (\cite{yue2022toward,roy2022svd}) and loss re-weighting techniques (\cite{shirokikh2020universal}) have also demonstrated promise in mitigating imbalance. However, only a limited number of methods have specifically addressed the challenges posed by the long-tailed distribution of disease classes. Recently, novel approaches have emerged to address long-tailed learning for medical image classification. Balanced-MixUp (\cite{galdran2021balanced}) addresses long-tailed distributions by mixing instance-based and class-based samples within the dataset. Another method groups rare classes into subsets based on prior knowledge and employs knowledge distillation to learn these subsets (\cite{ju2021relational}). Additionally, \cite{zhang2021mbnm} proposed resampling tail classes using a memory module combined with a re-weighting loss function to enhance long-tailed classification performance. Furthermore, a robust asymmetric loss function has been introduced, which regularizes the loss using the Hill loss approach for multi-label long-tailed learning (\cite{park2023robust}).\\
Despite these advancements, current methods still struggle with the accurate classification of long-tailed disease distributions. While some recent approaches attempt to address this issue through modified loss functions, they often overlook the potential of radiologist-guided disease pattern learning. For instance, integrating eye-gaze data could help deep learning models focus on critical regions, particularly for rare diseases. In this work, we propose a novel approach that incorporates radiologist eye-gaze patterns to improve long-tailed thoracic disease classification.

\begin{figure*}[h]
    \centering
    \includegraphics[width=1\linewidth]{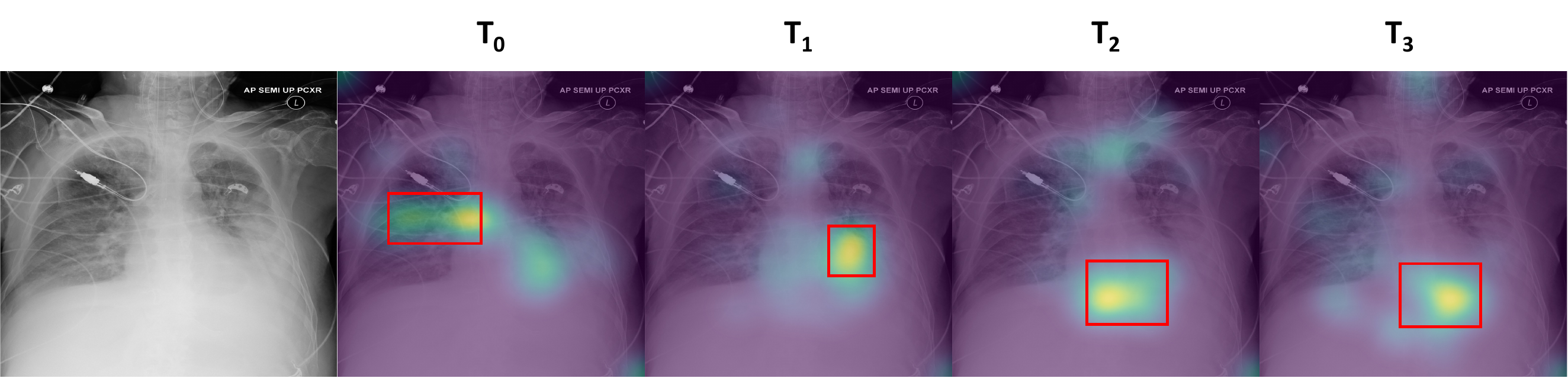} 
    \caption{Radiologists’ interpretation: Heatmap visualization of model attention across four timepoints ($T_0$, $T_1$, $T_2$, $T_3$) on chest X-ray  images. The red boxes highlight regions of interest as identified by the model. The focus shifts from areas associated with  atelectasis, classified as the 'head' class, at $T_0$ and $T_1$, to areas related to a hiatal hernia, classified as the 'tail' class, at $T_2$ and $T_3$.}
\label{fig:radint}
\end{figure*}
\section{Method}
\begin{algorithm}[h]
\caption{Generation of Visual Attention Heatmaps from Eye-Gaze Fixations}
\label{alg:heatmap_generation}
\begin{algorithmic}[1]

\REQUIRE Chest X-ray image $I$, gaze fixation points $G = \{(x_i,y_i)\}_{i=1}^{N}$, chest bounding box $B=(x_{min},y_{min},x_{max},y_{max})$

\ENSURE Visual attention heatmap $H$

\STATE Load image $I$ and obtain image dimensions $(H_I, W_I)$

\STATE Initialize empty fixation set $F$

\FOR{each fixation point $(x_i,y_i)$ in $G$}
    \IF{$x_{min} \leq x_i \leq x_{max}$ AND $y_{min} \leq y_i \leq y_{max}$}
        \STATE Add $(x_i,y_i)$ to $F$
    \ENDIF
\ENDFOR

\STATE Append boundary points $(0,0)$ and $(W_I,H_I)$ to $F$

\STATE \textbf{Global Attention:}
\STATE Generate 2D histogram $H_g$ from fixation coordinates in $F$

\STATE Apply Gaussian smoothing:
\[
H_g = \text{GaussianFilter}(H_g, \sigma_g)
\]

\STATE \textbf{Focal Attention:}
\STATE Cluster fixation points using K-means ($k=2$)

\STATE Select dominant fixation cluster $C$

\STATE Generate histogram $H_f$ from cluster $C$

\STATE Apply Gaussian smoothing:
\[
H_f = \text{GaussianFilter}(H_f, \sigma_f)
\]

\STATE Overlay heatmap on chest X-ray image

\RETURN Heatmap visualization $H$

\end{algorithmic}
\end{algorithm}
The proposed method is shown in Figure~\ref{fig:main}. We implement a teacher-student-based knowledge distillation (KD) framework in which the teacher is trained with snapshots of radiologist's eye gaze patterns at different time points called temporal-Human Visual Attention (t-HVA). During inference, we freeze the pre-trained teacher and only train the student for long-tailed thoracic disease classification. In Section \ref{tw}, we discuss the Time Windowed Integration and Disintegration blocks along with the loss functions. In Section \ref{training}, we discuss how the feature maps of t-HVA are calculated and how the teacher is trained. Finally, in Section \ref{kd}, we discuss the KD from the t-HVA teacher to the student for long-tailed classification.\\
\textbf{Preliminary.}
For computing t-HVA, we use the CXR image $\mathcal{I}\in\mathbb{R}^{H\times W\times C}$, where $H$ is the height, $W$ is the width, and $C$ is the number of channels of the image, and the corresponding radiologist eye gaze pattern $\mathbb{G}\in\mathbb{R}^{(\mathcal{F}, \mathcal{T})}$. Here, $\mathcal{F}$ are the eye gaze fixations over time $\mathcal{T}$ collected for CXR image $\mathcal{I}$. During inference, we only have a CXR image $\hat{\mathcal{I}}\in\mathbb{R}^{H\times W \times C}$ for which we need to predict a label $y$. Here, the objective is a long-tailed classification task where $y$ constitutes the  \textit{head}, \textit{medium}, and \textit{tail} classes.
\subsection{Time Windowed Integration-Disintegration}
\label{tw}
To calculate integration attention, we define a threshold distance between fixation points and select all points within this distance to designate as the main attention region. All other points are assigned to substitute attention regions. Together, the main and substitute attention regions form the integration attention. Gaussian filtering with a sigma value of 64 is then applied to these regions to generate the integration attention heatmap. In contrast, for disintegration attention, Gaussian filtering is applied directly to the raw fixation points with a sigma value of 128, producing the disintegration attention heatmap (Figure \ref{fig:eyegaze}.A). The duration of a radiologist's image viewing, referred to as fixation duration, is divided into n equal time partitions (in our case n=4). For instance, if the total fixation duration is 60 seconds, each partition would span 15 seconds. The fixation points are then distributed across these four time windows. For example, if there are 100 fixation points in total, there might be 20 points in the first 15-second window, 40 points in the next 15-second window, and so forth (Figure \ref{fig:eyegaze}.B). Within each time window, integration and disintegration attention are calculated using the corresponding fixation points. This integration (I) and disintegration (D) attention calculation is shown in Figure \ref{fig:eyegaze}.C. The integration attention is represented as $f^I$ and the disintegration attention is representation as $f^D$. For different time windows, the integration attention HVA maps and the disintegration attention HVA maps are shown in  Figure \ref{fig:radint}, respectively. 

\subsection{Training}
\label{training}
We train our proposed model using a teacher-student approach (Figure \ref{fig:architecture}). In this framework, the teacher block is trained using radiologists' eye gaze patterns, while the student block is trained for long-tailed classification. The teacher block consists of Time Windowed Integration (TW-I) and Time Windowed Disintegration (TW-D) blocks. The TW-I block learns to aggregate temporally distributed visual attention maps from the eye-gaze fixation points by computing an integration attention loss between its attention maps and those attention of the intermediate layers from the teacher model, which processes full temporal sequences. This encourages the TW-I block to mimic how radiologists integrate gaze information over time to arrive at diagnostic decisions. Conversely, the TW-D block learns to decompose global diagnostic cues into temporally localized attention maps, guided by a disintegration attention loss that aligns its outputs with the temporal structure of the teacher's attention. This dual mechanism ensures that the student model can both synthesize and localize relevant temporal features, enabling more robust learning under long-tailed label distributions. The TW-I block contains four sub-blocks, each utilizing Focal Attention to compute attention. Focal Attention computes attention using the focal modulation mechanism and hence attention for each TW-I block is computed sequentially (Figure \ref{fig:architecture}.A). Similarly, the TW-D block also comprises four sub-blocks, with each layer computing attention using global context attention (Figure \ref{fig:architecture}.B). 
The CXR image is fed as input to both the TW-I and TW-D blocks, where each block processes the image through four sequential attention layers. The output of the $t$-th layer in the TW-I block is denoted as $\mathcal{O}^I_t$, and similarly, the output of the $t$-th layer in the TW-D block is denoted as $\mathcal{O}^D_t$, for $t \in {1, 2, 3, 4}$. The TW-I and TW-D blocks are trained with their respective visual attention losses, $\mathcal{L}_{I\text{-}tVAL}$ and $\mathcal{L}_{D\text{-}tVAL}$, which enforce alignment between the predicted attention maps and radiologists’ gaze-derived attention, guiding each block to model temporal integration and disintegration of diagnostic cues effectively. This loss function minimizes the distance between the intermediate layers of the TW blocks and the ground truth human visual attention (HVA) windows (Figure \ref{fig:architecture}.C). 
\begin{equation}
\label{loss_equation_1}
    \mathcal{L}_{I-tVAL}=\sum_{i=1}^L\norm{ \frac{\mathcal{O}_t^I}{\norm{\mathcal{O}_t^I}}_2 - \frac{f_t^I}{\norm{f_t^I}}_2}_2
\end{equation}
\begin{equation}
\label{loss_equation_2}
    \mathcal{L}_{D-tVAL}=\sum_{i=1}^L\norm{ \frac{\mathcal{O}_t^D}{\norm{\mathcal{O}_t^D}}_2 - \frac{f_t^D}{\norm{f_t^D}}_2}_2
\end{equation}
where $\mathcal{O}^I$ is the output of the TW-I block and $\mathcal{O}^D$ is the output of the TW-D block.
\subsection{Knowledge Distillation}
\label{kd}
For long-tailed classification, we distill knowledge from a teacher model, pretrained with radiologists' eye gaze data, to a student model. The teacher-Human Visual Attention model (t-HVA) comprises the Time Windowed Integration (TW-I) and Time Windowed Disintegration (TW-D) blocks, which together encode both global and localized temporal gaze patterns of radiologists. The student model, in contrast, adopts a standard ResNet architecture. During training, the pretrained teacher's weights are frozen and used to guide the student via feature-level supervision. Specifically, the fused representations from TW-I and TW-D are transferred to the student through knowledge distillation. This setup ensures that the student not only learns discriminative features from limited data but also inherits the diagnostic priors embedded in expert gaze patterns, while being trained with a long-tailed loss tailored to handle class imbalance. The CXR image is simultaneously input to both the student block and the frozen teacher block, producing feature representations $f_s$ and $\{f_I, f_D\}$, respectively. Here, $f_I$ is the feature from the frozen TW-I and $f_D$ is the feature from the frozen TW-D. The knowledge distillation loss facilitates this process by transferring the joint representation from the teacher block ($J(f_I, f_D)$) to the student block ($f_s$), minimizing the Bhattacharyya distance between their probability distributions. The feature maps extracted from the teacher and student networks are first normalized using a softmax operation to obtain probability distributions over spatial locations. These normalized distributions are then used to compute the Bhattacharyya distance. The final student model training incorporates a combination of a Label-Distribution-Aware Margin (LDAM) loss ($\mathcal{L}_{LDAM}$) and knowledge distillation loss ($\mathcal{L}_{BD}$), ensuring effective learning from the teacher model while addressing class imbalance.
\begin{equation}
    \mathcal{L}_{BD}=-ln\bigg(\int\sqrt{f_s*J(f_I, f_D)}\bigg)
\end{equation}
\begin{equation}
    \mathcal{L}_{LDAM}=-\log\frac{e^{z_y-\Delta_y}}{e^{z_y-\Delta_y}+\sum_{j\neq y}e^{z_j}} 
\end{equation}
where $z_y$ is the score of the class $y$ and $\Delta_y$ is the margin for the class $y$. Finally, the student loss, $\mathcal{L}_s$ is defined as $\mathcal{L}_s = \mathcal{L}_{LDAM}+\lambda\mathcal{L}_{BD}$.

\section{Implementation}
\subsection{Dataset Curation}
We utilize eye gaze fixation data from the EGD-CXR and REFLACX datasets. The EGD-CXR dataset contains eye gaze data from one radiologist, whereas the REFLACX dataset contains multiple radiologists' eye gaze data. The corresponding chest X-ray (CXR) images for these eye gaze datasets are obtained from the MIMIC-CXR-JPG dataset. The eye gaze data is used for training the teacher model. For long-tailed classification experiments, we employ the publicly available NIH-CXR-LT and MIMIC-CXR-LT datasets. We adhere to the standard training, balanced testing, and test splits (\cite{holste2022long}) (Details in Supplementary Figure \ref{fig:supp1}).

\subsection{Model Hyperparameters}
Fixation points within the chest bounding box are clustered using K-means (Euclidean distance, 10 restarts), and the cluster with lower spatial variance is designated as focal attention. The implicit distance threshold ($D_k = \max_i |\mathbf{x}_i - \boldsymbol{\mu}_k|_2$) is determined by K-means without manual tuning. Fixation density maps are constructed via 2D histograms followed by Gaussian smoothing with $\sigma=128$ for integration and $\sigma=64$ for disintegration, selected through validation grid search over \{32, 64, 96, 128, 160\}. The global teacher is GC-ViT-Tiny (depths=[3,4,19,5], heads$=[2,4,8,16]$, dim=64, window sizes$=[7,7,14,7]$, MLP ratio=3, drop path=0.2), initialized with ImageNet-1K weights. The focal teacher is FocalNet-Tiny-SRF (depths =[2,2,6,2], dim=96), also ImageNet pretrained. Both teachers are adapted to single-channel CXR by averaging RGB weights in the first convolution and replacing the classification head. The student network is ResNet-50 (ImageNet pretrained), optimized with SGD (momentum=0.9, weight decay=1e-4, initial LR=1e-3 with cosine decay, batch size= 32), with distillation applied at Layer3 and Layer4 using $1\times1$ feature projection before the Bhattacharyya loss.

\subsection{Evaluation Metrics}
We report different metrics based on the test set distribution and task type. For the balanced test set, we report average accuracy (↑), computed as the unweighted mean of per-class accuracies. In addition to the overall average accuracy, we report the group-wise average accuracy, calculated as the mean of the accuracies for the head, medium, and tail class groups. This metric captures how well the model performs across different class frequency regimes.\\
For the imbalanced test set, which reflects the real-world long-tailed distribution, we report balanced accuracy (↑), which averages recall across all classes and reduces the bias toward frequent classes.\\
For multiclass classification experiments, we report four metrics: balanced accuracy (↑), Matthews Correlation Coefficient (MCC) (↑), Area Under the ROC Curve (AUC) (↑), and Weighted F1 Score (wF1) (↑). These metrics collectively provide a comprehensive evaluation of model performance across balanced, imbalanced, and multiclass settings.

\subsection{Experiments}
The time windows are computed by splitting the entire duration of the eye gaze collection process into $4$ equal windows. These splits are done based on time duration. For TW-I training, we use a learning rate ($lr$) of $1e-4$ and train for $100$ epochs. For TW-D training, we use a $lr$ of $5e-4$ and train for $250$ epochs.
For both, we use Adam optimizer with StepLR scheduling (step size=10. $\gamma$
=0.1). For student training, we train for 100 epochs with a $lr$ of $1e-4$ and Adam optimizer. We train on a single Quadro RTX 8000 (48 GB) with a batch size of 256.

\begin{table*}[t]
\small
\centering
\caption{\textbf{Long-Tailed Classification.} The results are shown for NIH-CXR-LT and MIMIC-CXR-LT datasets. We report average accuracy ($\uparrow$) for the balanced test set and balanced accuracy ($\uparrow$) for the imbalanced test set. The best and the second best results are shown in \textcolor{mycoral}{coral} and \textcolor{mylightblue}{light blue} respectively.}\label{tab1}
\begin{tabular}{ccccccccccc}
\hline
Datasets($\rightarrow$) & \multicolumn{5}{c}{\bfseries NIH-CXR-LT} & \multicolumn{5}{c}{\bfseries MIMIC-CXR-LT}\\
\hline
Baselines($\downarrow$) & \multicolumn{4}{c}{Balanced test} & \multicolumn{1}{c}{Test} & \multicolumn{4}{c}{Balanced test} & \multicolumn{1}{c}{Test}\\
\hline
 & Head & Medium & Tail & Avg & bAcc & Head & Medium & Tail & Avg & bAcc\\
\hline
\hline
Softmax & 0.419 & 0.056 & 0.017 & 0.164 & 0.115 & 0.503 & 0.039 & 0.022 & 0.188 & 0.169\\
CB Softmax & 0.295 & 0.415 & 0.217 & 0.309 & 0.269 & 0.493 & 0.167 & 0.222 & 0.294 & 0.227\\
RW Softmax & 0.248 & 0.359 & 0.258 & 0.288 & 0.260 & 0.473 & 0.139 & 0.133 & 0.249 & 0.211\\
Focal Loss & 0.362 & 0.056 & 0.042 & 0.153 & 0.122 & 0.477 & 0.044 & 0.022 & 0.181 & 0.172\\
CB Focal Loss & 0.371 & 0.333 & 0.117 & 0.274 & 0.232 & 0.373 & 0.117 & 0.344 & 0.278 & 0.191\\
RW Focal Loss & 0.286 & 0.293 & 0.117 & 0.232 & 0.197 & 0.403 & 0.283 & 0.211 & 0.299 & 0.239\\
LDAM & 0.410 & 0.133 & 0.142 & 0.228 & 0.178 & 0.497 & 0.000 & 0.000 & 0.166 & 0.165\\
CB LDAM & 0.357 & 0.285 & 0.208 & 0.284 & 0.235 & 0.467 & 0.161 & 0.211 & 0.280 & 0.225\\
CB LDAM-DRW & 0.476 & 0.356 & 0.250 & 0.361 & 0.281 & 0.520 & 0.156 & 0.356 & 0.344 & 0.267\\
RW LDAM & 0.305 & 0.419 & 0.292 & 0.338 & 0.279 & 0.437 & 0.250 & 0.167 & 0.284 & 0.243\\
RW LDAM-DRW & 0.410 & 0.367 & \cellcolor[HTML]{85C1E9} 0.308 & \cellcolor[HTML]{85C1E9} 0.362 & 0.289 & 0.447 & 0.256 & 0.311 & 0.338 & 0.275\\
MixUp & 0.419 & 0.044 & 0.017 & 0.160 & 0.118 & 0.543 & 0.011 & 0.011 & 0.189 & 0.176\\
Balanced-MixUp & 0.443 & 0.081 & 0.108 & 0.211 & 0.155 & 0.480 & 0.039 & 0.011 & 0.177 & 0.168\\
Decoupling-cRT & 0.433 & 0.374 & 0.300 & 0.369 & \cellcolor[HTML]{85C1E9} 0.294 & 0.490 & 0.306 & \cellcolor[HTML]{85C1E9} 0.367 & \cellcolor[HTML]{85C1E9} 0.387 & \cellcolor[HTML]{EC7063} 0.296\\
Decoupling-$\tau$-norm & 0.457 & 0.230 & 0.083 & 0.257 & 0.214 & 0.520 & 0.167 & 0.067 & 0.251 & 0.230\\
\hline
GazeRadar & 0.390 & 0.074 & 0.075 & 0.187 & 0.140 & 0.527 & 0.006 & 0.000 & 0.279 & 0.174\\
RadioTransformer & 0.386 & 0.093 & 0.083 & 0.193 & 0.131 & 0.493 & 0.0 & 0.0 & 0.260 & 0.164\\
\hline
\hline
GazeLT (Ours) & 0.404 & 0.411 & \cellcolor[HTML]{EC7063} 0.417 & \cellcolor[HTML]{EC7063} 0.41 & \cellcolor[HTML]{EC7063} 0.315 & 0.48 & 0.278 & \cellcolor[HTML]{EC7063} 0.489 & \cellcolor[HTML]{EC7063} 0.418 & \cellcolor[HTML]{85C1E9} 0.292 \\
\hline
\label{tab:tab1}
\end{tabular}
\end{table*}
\subsection{Statistical Analysis}
We employed a comprehensive set of statistical methods to rigorously evaluate the performance of our approach. For long-tailed classification analysis, we used balanced accuracy to account for class imbalance on the balanced test set, and both F1 score and accuracy to assess overall performance on the test set. To determine the statistical significance of differences in performance metrics, we conducted t-tests, ensuring that observed differences were not due to random chance. Additionally, we utilized Optuna for hyperparameter optimization, leveraging its advanced algorithms to efficiently search for the best hyperparameter values and enhance model performance. For the classification experiments, we report the Matthews Correlation Coefficient (MCC), Area under the ROC Curve (AUC), and weighted F1 (wF1) scores.
\section{Results}
\subsection{Patient Characteristics} 
We employ the NIH-CXR-LT and MIMIC-CXR-LT datasets for our long-tail classification task, which includes various thoracic disease pathologies alongside normal images. Based on the frequency of cases in the training set, the datasets are organized into three categories: head (more than 1,000 patients), medium (100–1,000 patients), and tail (fewer than 100 patients). In the NIH-CXR-LT dataset, the tail classes are Pneumoperitoneum, Hernia, Subcutaneous Emphysema, and Pneumomediastinum. Meanwhile, the MIMIC-CXR-LT dataset's tail classes include Pneumoperitoneum, Subcutaneous Emphysema, and Pneumomediastinum. The average age of patients in the NIH-CXR-LT training set is 46.13 ± 16.46 years, with 37,872 male patients and 30,186 female patients (Supplementary Table \ref{tab:supp1}-\ref{tab:supp3}).
\begin{table}[ht]
\centering
\caption{\textbf{Ablation results.} We compare different components of the \textit{GazeLT} architecture. We report the average accuracy ($\uparrow$) for the balanced set and balanced accuracy ($\uparrow$) for the imbalanced test set of NIH-CXR-LT dataset.}\label{tab1}
\begin{tabular}{ccc}
\hline
Components($\downarrow$) & Balanced set & Test set\\
\hline
\hline
Integration (I) & 0.383 & 0.296 \\
Disintegration (D) & 0.392 & 0.294\\
TW-I & \cellcolor[HTML]{85C1E9} 0.397 & \cellcolor[HTML]{85C1E9} 0.3\\
TW-D & 0.377 & 0.299\\
\hline
I+D & 0.363 & 0.295\\
TW-I+TW-D & \cellcolor[HTML]{EC7063} 0.41 & \cellcolor[HTML]{EC7063} 0.315\\
\hline
\label{tab:ablation_nih}
\end{tabular}
\end{table}

\begin{table}[t]
\centering
\caption{\textbf{Ablation results (MIMIC-CXR-LT).} We compare different components of the \textit{GazeLT} architecture. We report the average accuracy ($\uparrow$) for the balanced set and balanced accuracy ($\uparrow$) for the imbalanced test set of MIMIC-CXR-LT dataset.}\label{tab1}
\begin{tabular}{ccc}
\hline
Components($\downarrow$) & Balanced set & Test set\\
\hline
\hline
TW-I & \cellcolor[HTML]{85C1E9} 0.375 & \cellcolor[HTML]{85C1E9} 0.275\\
TW-D & 0.37 & 0.254\\
\hline
TW-I+TW-D & \cellcolor[HTML]{EC7063} 0.418 & \cellcolor[HTML]{EC7063} 0.292\\
\hline
\label{tab:ablation_mimic}
\end{tabular}
\end{table}
\subsection{Long-tailed Classification} 
Here, we present the long-tailed classification results of GazeLT. Our results are compared with models trained using standard loss functions like Softmax and Focal, other standard LT baselines like Label-Distribution-Aware Margin (LDAM), re-weighting LDAM with optional deferred re-weighting (RW LDAM-DRW, and Decoupling- classifier re-training (Decoupling-cRT) (\cite{cao2019learning,cui2019class}) and visual attention (VA) pre-trained baselines i.e. RadioTransformer and GazeRadar. From Table \ref{tab:tab1}, we observe that, on the NIH-CXR-LT balanced set, our proposed method GazeLT outperforms LDAM by 18.2\%, RW LDAM-DRW by 4.8\%, and Decoupling-cRT by 4.1\%. Whereas, GazeLT significantly outperforms the VA baselines RadioTransformer by 21.7\% (p$<$0.001) and GazeRadar by 22.3\%  (p$<$0.001). Similar performance is achieved for the NIH-CXR-LT test set with GazeLT outperforming LDAM by 13.7\%, RW LDAM-DRW by 2.6\%, and Decoupling-cRT by 2.1\%, and significantly outperforming the VA baselines RadioTransformer and GazeRadar by $>$15\%. Similarly, for MIMIC-CXR-LT, GazeLT outperforms all baselines on the balanced set and achieves a balanced accuracy of 0.292 where the best baseline (Decoupling-cRT) achieves 0.296. It is important to note here that the proposed method It is important to note that the proposed method significantly outperforms the baseline methods on the tail classes, achieving improvements of 10.9\% on the NIH-CXR-LT dataset and 12.2\% on the MIMIC-CXR-LT dataset (p $<$ 0.001). 
While GazeLT shows a slight 5\% drop in head-class accuracy, it achieves significant gains in tail-class performance, which are often more clinically critical. Overall, GazeLT attains the highest average accuracy across both datasets: 0.410 on NIH-CXR-LT (13.26\% improvement over the best baseline) and 0.418 on MIMIC-CXR-LT (8.01\% improvement). These results highlight the model's improved diagnostic coverage across both common and rare conditions. Additional results on the individual medium classes are shown in Figure \ref{fig:results}.A and the individual tail classes are shown in Figure \ref{fig:results}.B. 
\begin{table}
\centering
\caption{\textbf{Multiclass classification.} We compare GazeLT with Gazeradar and RadioTransformer. We report the balanced accuracy(↑), MCC(↑), AUC(↑) and wF1(↑) 
for the imbalanced test set of NIH-CXR-LT dataset.
.}\label{tab1}
\begin{tabular}{ccccc}
\hline
NIH-CXR-LT & bAcc & MCC & AUC & wF1\\
\hline
\hline
GazeRadar &  0.187 & 0.175 & 0.791 & 0.17\\
RadioTransformer & 0.193 & 0.197 & 0.79 & 0.183\\
I+D & \cellcolor[HTML]{85C1E9} 0.363 & \cellcolor[HTML]{85C1E9} 0.337 & \cellcolor[HTML]{85C1E9} 0.821 & \cellcolor[HTML]{85C1E9} 0.316\\
\hline
TW-I+TW-D & \cellcolor[HTML]{EC7063} 0.421 & \cellcolor[HTML]{EC7063} 0.391 & \cellcolor[HTML]{EC7063} 0.829 & \cellcolor[HTML]{EC7063} 0.394\\
\hline
\label{tab:tab3}
\end{tabular}
\end{table}
\subsection{Ablation}
We ablate different components of GazeLT (Table \ref{tab:ablation_nih}). The teacher of GazeLT is trained using the time windowed HVAs which is a combination of TW-I and TW-D. We show that on the NIH-CXR-LT balanced test set, the balanced accuracy for the TW-I-only block is 0.397, and for the TW-D-only block is 0.377. Also, the balanced accuracy for the I-only block is 0.383, the D-only block is 0.392 and the combined I and D blocks is 0.363. Our proposed method, GazeLT (TW-I+TW-D) outperforms the combined I and D by 4.7\%. This result validates our hypothesis that time window-based attention calculation improves long-tailed disease classification. We also show the ablations results on the MIMIC-CXR-LT dataset in Table~\ref{tab:ablation_mimic}.

\subsection{Multi-class Classification}
\begin{figure*}[h]
    \centering
    \includegraphics[width=1\linewidth]{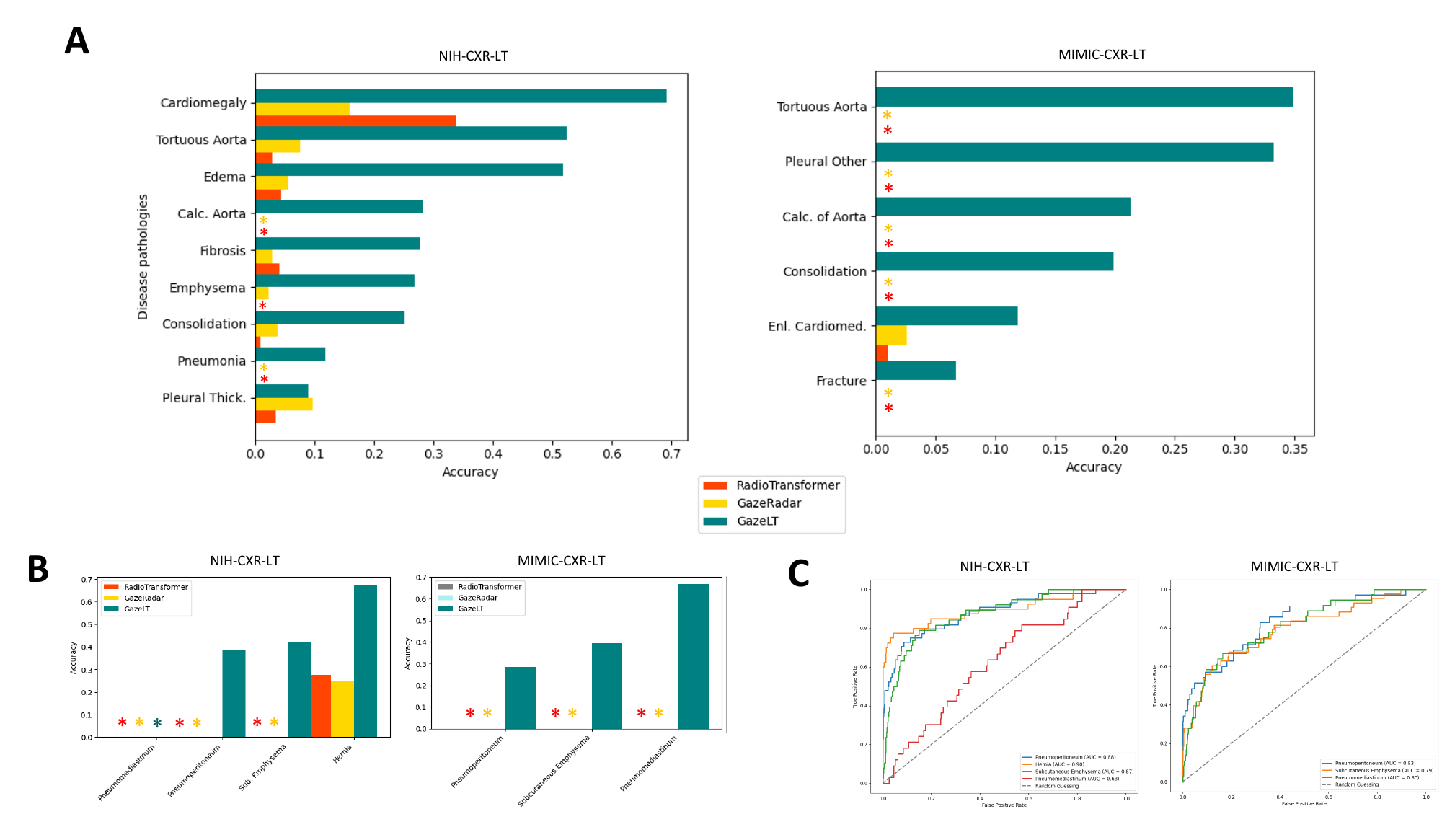} 
    \caption{A. Accuracy of medium classes on NIH-CXR-LT and MIMIC-CXR-LT datasets, comparing GazeLT, GazeRadar, and RadioTransformer, B. Accuracy of tail classes on NIH-CXR-LT and MIMIC-CXR-LT datasets, comparing GazeLT, GazeRadar, and RadioTransformer (* represents 0), C. ROC-AUC curves for tail classes on NIH-CXR-LT and MIMIC-CXR-LT datasets.
}
\label{fig:results}
\end{figure*}

For the NIH-CXR-LT dataset, our proposed method, GazeLT, achieves a balanced accuracy score of 0.421, outperforming GazeRadar (0.187) and RadioTransformer (0.193). In the multi-class classification task, GazeLT attains an AUC of 0.829, significantly exceeding the performance of GazeRadar (AUC: 0.791) and RadioTransformer (AUC: 0.790) (Table \ref{tab:tab3}). The ROC curves for the tail classes are presented in Figure \ref{fig:results}.C. Notably, GazeLT with TW-I+TW-D outperforms the I+D approach across all classification metrics. TW-I+TW-D achieves an MCC of 0.392, an AUC of 0.829, and a weighted F1 score of 0.394, surpassing the I+D scores of 0.337, 0.821, and 0.316, respectively.
\begin{table}[t]
\centering
\caption{\textbf{Different windows on the balanced set (NIH-CXR-LT).}}\label{tab1}
\begin{tabular}{ccccc}
\hline
Windows($\downarrow$) & Head & Medium & Tail & Avg\\
\hline
\hline
2 windows & 0.348 & \cellcolor[HTML]{85C1E9} 0.396 & \cellcolor[HTML]{85C1E9} 0.408 & 0.382\\
4 windows (Ours) & \cellcolor[HTML]{85C1E9} 0.404 & \cellcolor[HTML]{EC7063} 0.411 & \cellcolor[HTML]{EC7063} 0.417 & \cellcolor[HTML]{EC7063} 0.41\\
8 windows & \cellcolor[HTML]{EC7063} 0.486 & \cellcolor[HTML]{85C1E9} 0.396 & 0.275 & \cellcolor[HTML]{85C1E9} 0.403\\
\hline
\label{tab:time_window}
\end{tabular}
\end{table}
\subsection{Sensitivity Analysis of Time Window Parameter}
To quantitatively evaluate the impact of temporal granularity on model performance, we conducted a sensitivity analysis by varying the number of time windows $n \in {2, 4, 8}$. We report results on the balanced test set, focusing on average accuracy as well as class-group performance across head, medium, and tail categories. As summarized in Table~\ref{tab:time_window}, the model with $n=4$ achieved the highest average accuracy (0.410), compared to 0.382 and 0.403 for $n=2$ and $n=8$, respectively. These results indicate that the 4-window configuration offers an effective trade-off between capturing sufficient temporal detail and avoiding overfitting.\\
To complement the quantitative findings, we visualize raw gaze points overlaid on chest X-ray images for a representative case in Rebuttal Figure~\ref{fig:248_0}. This example includes Atelectasis (head class) and Hiatal Hernia (tail class), marked with green and red bounding boxes, respectively (Another example in Rebuttal Figure~\ref{fig:248_1}). With $n=2$, the gaze points are broadly distributed across both windows, resulting in no window distinctly capturing attention over either the head or tail class. With $n=8$, we observe redundant attention—multiple windows contain overlapping gaze regions that cover both head and tail classes, reducing the temporal disentanglement needed for focused representation learning. In contrast, the $n=4$ configuration demonstrates a more structured allocation of gaze, where one window primarily aligns with the head class and another aligns with the tail class. This separation enables more effective attention modeling and class-specific representation learning.\\
Based on these observations, we adopt $n=4$ in all experiments throughout the paper, as it provides optimal performance while preserving interpretability and reducing redundancy in temporal attention patterns.
\section{Discussions}

This study introduces GazeLT, a novel deep learning framework that incorporates the temporal visual attention patterns of radiologists to improve long-tailed classification in medical imaging. Long-tailed classification is challenging due to the imbalance between common (head) and rare (tail) disease classes in medical datasets. GazeLT addresses this challenge by using a teacher-student model where the teacher is trained with radiologists' eye gaze patterns and the student model benefits from this expertise for better diagnostic accuracy.

\noindent The primary innovation of GazeLT is its focus on the temporal dynamics of visual attention. Traditional models typically use static aggregated attention maps, but this study decomposes visual attention into time-windowed segments, capturing the dynamic nature of radiologists' gaze patterns. This temporal resolution allows the DL model to replicate the sequential and hierarchical nature of visual search patterns, leading to better performance in disease classification.\\
Incorporating clinician-in-the-loop strategies significantly enhances the interpretability of DL models. Radiologists' eye gaze patterns provide clinically relevant cues that guide the model's focus towards important regions in the medical images. This alignment with human experts' decision-making processes not only improves diagnostic accuracy but also offers a more transparent rationale for the model's predictions. This is crucial as DL models in radiology often learn shortcuts or focus on irrelevant features, leading to potential misdiagnoses.\\
Previous works on long-tailed learning have focused on mitigating class imbalance through methods like reweighting, class balancing, and decoupling (\cite{cao2019learning,cui2019class}) often relying on modified loss functions. In contrast, this study introduces a novel teacher-student framework that integrates radiologists' gaze-based visual attention patterns to address long-tailed disease classification. We retain the original loss function and enhance learning by aligning model attention with expert cues. This is the first work to leverage human visual attention in a time-windowed paradigm, offering a unique perspective for improving model performance in imbalanced medical datasets.\\
The teacher-student framework of GazeLT ensures that the student model inherits the nuanced understanding of rare disease patterns from the teacher model, which is trained with the time-windowed visual attention data. Experiments on both datasets demonstrate that GazeLT consistently outperforms baseline models, particularly in identifying tail classes. The comparative analysis shows significant improvements in balanced accuracy and F1 scores across both head and tail classes, outperforming models like RadioTransformer and GazeRadar that lack the temporal dynamics and comprehensive integration strategies of GazeLT.\\
Ablation studies further validate the contributions of the individual components of GazeLT. The superior performance of the combined TW-I and TW-D blocks over their standalone counterparts emphasizes the importance of both fine-grained and coarse attention patterns. The significant performance drop observed when these components are excluded or simplified further confirms that the integration of both attention strategies is essential for effective long-tailed classification.\\
A key strength of our proposed GazeLT architecture is its efficient use of radiologists’ eye gaze data, which is required only during the training of the teacher model. During inference, the model relies solely on the input image, significantly reducing computational overhead and precluding the need for additional eye gaze data accrual. Furthermore, unlike previous approaches that focus on single time-point attention, our method captures detailed, time-windowed gaze information, enabling a more comprehensive understanding of visual attention dynamics.\\
While \textit{GazeLT} demonstrates strong performance in long-tailed chest X-ray classification, it 
has certain limitations. A central dependency of the proposed framework is the availability of radiologist eye-tracking data during training, which may pose practical challenges related to data acquisition, standardization, and scalability. The collection of high-quality gaze 
fixation points requires dedicated eye-tracking hardware, controlled viewing conditions, and participation from expert clinicians—factors that may limit widespread adoption, particularly across institutions with varying resources and infrastructure. Importantly, however, gaze data is only utilized during the training phase to supervise the teacher network. Once the knowledge distillation process is complete, the student model operates independently and does not require gaze input at inference time. This design ensures that the deployment of \textit{GazeLT} remains feasible in clinical environments where eye-tracking data is unavailable.\\
Moreover, while our experiments focus exclusively on chest radiographs, the broader applicability of \textit{GazeLT} to other imaging modalities (e.g., CT, MRI, ultrasound) and clinical tasks (e.g., lesion detection, segmentation) remains an open question. In previous studies, eye gaze patterns collected from 3D brain MRIs have been leveraged for segmentation tasks. In our proposed framework, we demonstrate that eye gaze data acquired from 2D chest X-rays significantly improves performance on long-tailed classification. While our current work focuses on 2D data, the approach is readily extensible to other imaging modalities, including 3D scans and videos. Although collecting eye gaze data for 3D images poses practical challenges, prior works have already established feasibility by successfully acquiring such data. Hence, our method is inherently modality-agnostic and can be readily extended to diverse imaging modalities, including 2D, 3D, and dynamic (video) data. Additionally, the complexity of the teacher-student framework may increase computational requirements. Future work could explore the applicability  of GazeLT across different medical imaging modalities and the integration of multi-modal data to further enhance diagnostic accuracy and interpretability. Moreover, investigating the applicability of this approach to other domains with long-tailed distributions could provide broader insights into the potential of temporal visual attention in DL models.\\
In summary, GazeLT represents a significant advancement in medical imaging by integrating the temporal visual attention patterns of radiologists into a DL framework. This approach improves the interpretability and clinical relevance of model predictions and addresses the challenges of long-tailed classification. The study demonstrates the  importance of temporal dynamics in visual attention-based learning. As the field of medical AI evolves, incorporating expert knowledge into DL models will be crucial for developing robust, accurate, and interpretable diagnostic tools.

\section{Conclusion}
In this work, we present a visual attention-guided deep learning framework for long-tailed classification. Experts' temporal visual attention consists of context-rich \textit{integration} attention and coarse \textit{disintegration} attention. These attentions, captured at different time windows, when learned by transformer architectures, generate better visual attention-guided representations. At different time windows, the radiologists look at different disease patterns ranging from the \textit{head} classes to the \textit{tail} classes. These representations help in improving performance in long-tailed disease classification. One of the major limitations of our work is that we do not take into consideration the diagnosis confidence score of radiologists to compute the time windows. In the future, we aim to work on combining multiple radiologists' gaze patterns for long-tailed disease classification and evaluate the effect of changing the time windows.

\section*{Acknowledgements} This research was partially supported by National Institutes of Health (NIH) and National Cancer Institute (NCI) grants 1R21CA258493-01A1, 1R01CA297843-01, 3R21CA258493-02S1, 1R03DE033489-01A1, and National Science Foundation (NSF) grant 2442053. The content is solely the responsibility of the authors and does not necessarily represent the official views of the National Institutes of Health.

\section*{Data availability}
The datasets used in this study are publicly available. The EDG-CXR dataset is available at this \href{https://physionet.org/content/egd-cxr/1.0.0/}{PhysioNet link}, and the REFLACX dataset is available at this \href{https://physionet.org/content/reflacx-xray-localization/1.0.0/}{PhysioNet link}. The chest X-ray (CXR) images are obtained from the MIMIC-CXR-JPG dataset, which is available at the \href{https://physionet.org/content/mimic-cxr-jpg/2.1.0/}{PhysioNet link}. The long-tailed classification datasets, NIH-CXR-LT and MIMIC-CXR-LT, are available at this \href{https://github.com/vita-group/longtailcxr}{GitHub repository}.

\bibliography{references}

\clearpage
\appendix

\section{Inclusion and Exclusion criteria}
\begin{figure}[h]
    \centering
    \includegraphics[width=2\linewidth]{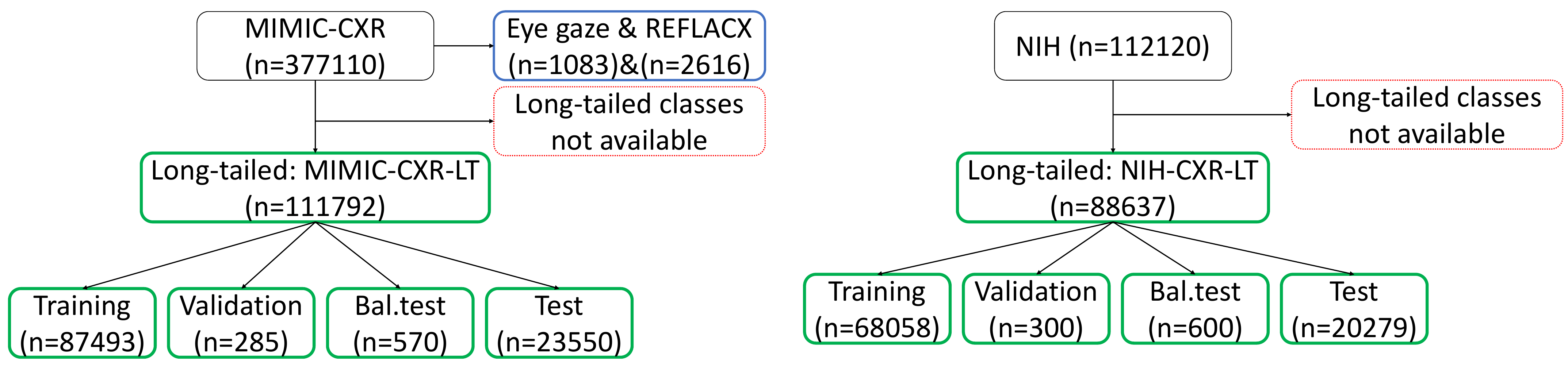} 
    \caption{Inclusion-Exclusion criteria.
}
\label{fig:supp1}
\end{figure}

\clearpage
\section{Patient demographics}
\begin{table}[h]
    \centering
    \begin{tabular}{c|c|c}
    Type&Class&n\\
    \hline
    \hline
         \multirow{7}{*}{Head} & No Finding & 45439 \\
         & Infiltration & 6941 \\
         & Atelectasis & 3135 \\
         & Effusion & 2875 \\
        & Nodule & 2036 \\
        & Mass & 1665 \\
        & Pneumothorax & 1485 \\
        \hline
        \multirow{9}{*}{Medium} & Consolidation & 868 \\
        & Pleural Thickening & 811 \\
        & Cardiomegaly & 800 \\
        & Fibrosis & 519 \\
        & Edema & 435 \\
        & Tortuous Aorta & 339 \\
        & Emphysema & 229 \\
        & Pneumonia & 213 \\
        & Calcification of the Aorta & 121 \\
        \hline
        \multirow{4}{*}{Tail} & Pneumoperitoneum & 59 \\
        & Hernia & 49 \\
        & Subcutaneous Emphysema & 32 \\
        & Pneumomediastinum & 7 \\
        \hline
    \end{tabular}
    \caption{NIH-CXR-LT}
    \label{tab:supp1}
\end{table}

\begin{table}[h]
    \centering
    \begin{tabular}{c|c|c}
    Type&Class&n\\
    \hline
    \hline
\multirow{10}{*}{Head} & No Finding & 53260 \\
& Lung Opacity & 7927 \\
& Cardiomegaly & 5113 \\
& Atelectasis & 4539 \\
& Pleural Effusion & 3832 \\
& Support Devices & 3279 \\
& Edema & 2395 \\
& Pneumonia & 2195 \\
& Pneumothorax & 1172 \\
& Lung Lesion & 1036 \\
\hline
\multirow{6}{*}{Medium} & Fracture & 791 \\
& Enlarged Cardiomediastinum & 638 \\
& Consolidation & 609 \\
& Pleural Other & 254 \\
& Calcification of the Aorta & 207 \\
& Tortuous Aorta & 175 \\
\hline
\multirow{3}{*}{Tail} & Pneumoperitoneum & 32 \\
& Subcutaneous Emphysema & 27 \\
& Pneumomediastinum & 12 \\
\hline
    \end{tabular}
    \caption{MIMIC-CXR-LT}
    \label{tab:supp2}
\end{table}

\begin{table}[h]
    \centering
    \begin{tabular}{c|c|c}
    Parameters&Categories&n\\
    \hline
    \hline
\multirow{2}{*}{Gender} & Male & 37872 \\
& Female & 30186 \\
\hline
\multirow{7}{*}{Age} & $<$20 & 4426 \\
& 20-30 & 7841 \\
& 30-40 & 10498 \\
& 50-60 & 16391 \\
& 70-80 & 3602 \\
& 80-90 & 529 \\
& 90-100 & 27 \\
\hline
    \end{tabular}
    \caption{Demographics}
    \label{tab:supp3}
\end{table}

\section{Additional Qualitative Results}
\begin{figure*}
    \centering
    \includegraphics[width=1\linewidth]{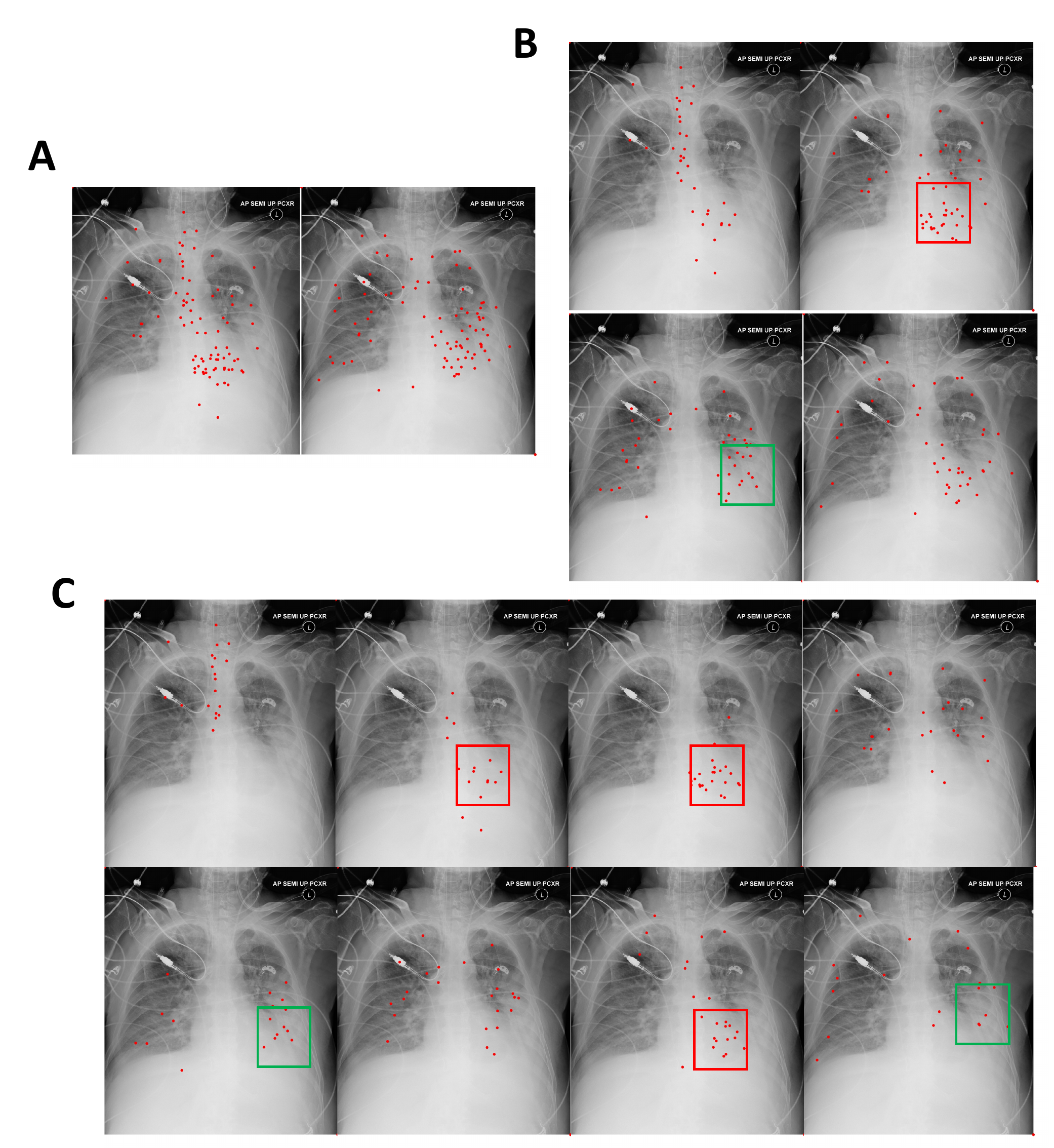}
    \caption{\textbf{Gaze point temporal alignment with head and tail class regions across different time window divisions.}
(A) Example of a patient with atelectasis (head class) and hiatal hernia (tail class). Gaze points are divided into two time windows, but neither window shows fixation on the relevant class regions.
(B) Gaze points divided into four time windows. This represents an ideal temporal separation: the head class region (\textcolor{red}{red} bounding box) and tail class region (\textcolor{green}{green} bounding box) are fixated in distinct time windows.
(C) Gaze points divided into eight time windows. Both head and tail class regions receive attention in multiple windows, indicating temporally overlapping focus across classes.}
    \label{fig:248_0}
\end{figure*}

\begin{figure*}
    \centering
    \includegraphics[width=1\linewidth]{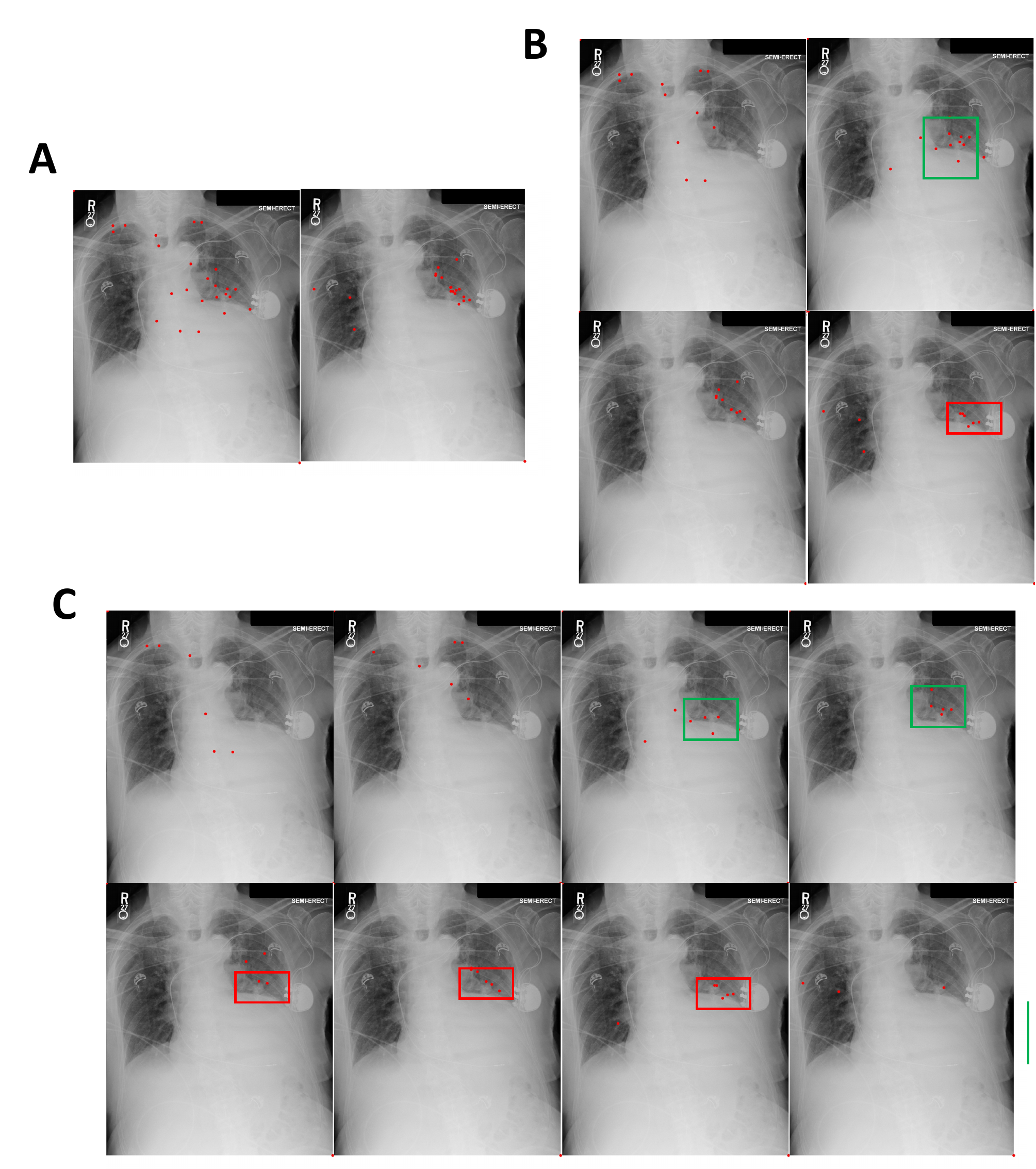}
    \caption{\textbf{Additional qualitative results.} Example of another patient with atelectasis (head class) and enlarged hilum (tail class).}
    \label{fig:248_1}
\end{figure*}

\begin{figure*}
    \centering
    \includegraphics[width=1\linewidth]{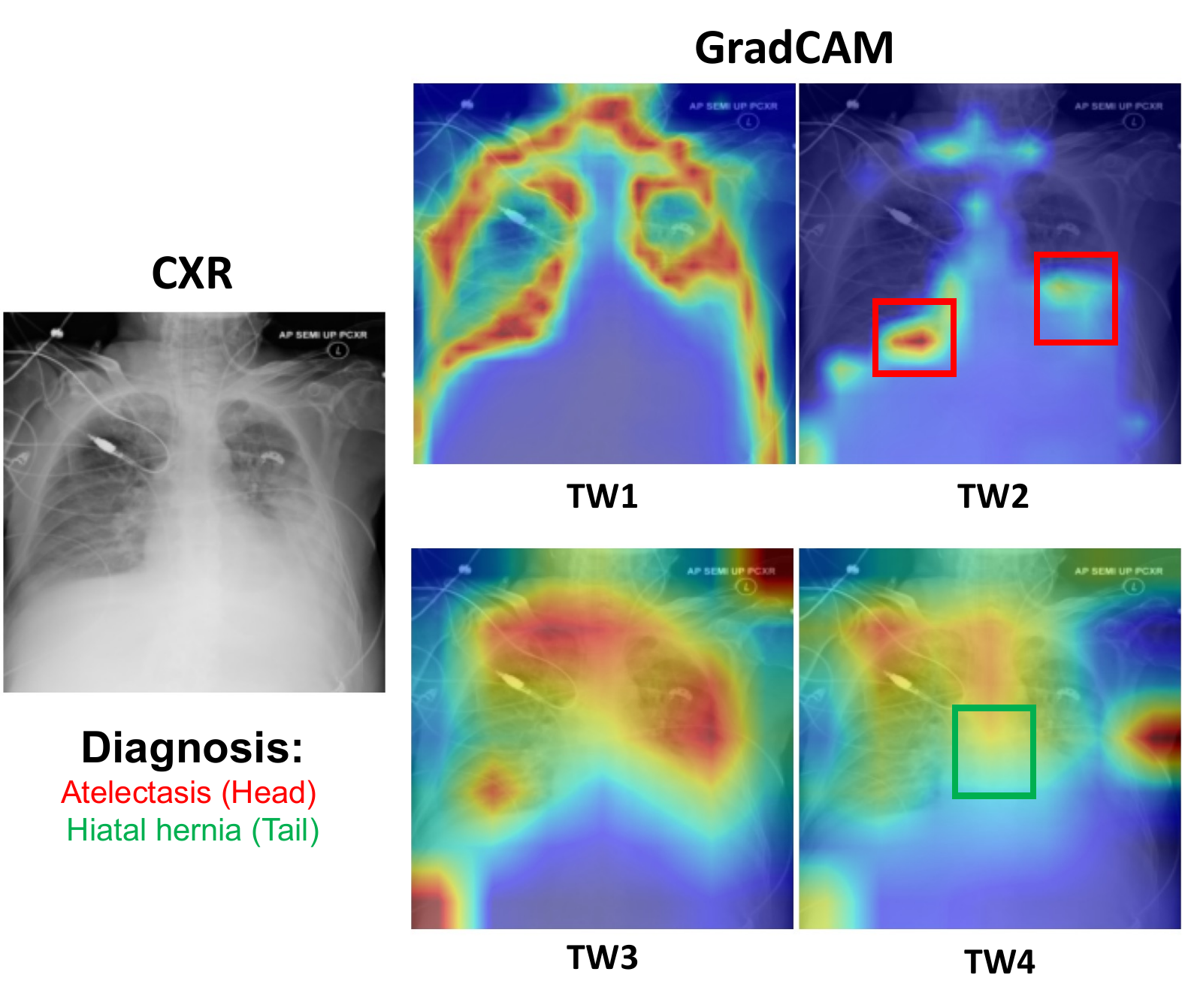}
    \caption{Visualization of gaze and model attention for a patient diagnosed with Atelectasis (head class) and Hiatal Hernia (tail class). GradCAM heatmaps from the model predictions are included for comparison. Radiologist-annotated regions corresponding to the head class are shown with red bounding boxes, and tail class regions with green bounding boxes.}
    \label{fig:gradcam}
\end{figure*}

\end{document}